\documentclass[letterpaper]{article} 
\usepackage{aaai2026}  
\usepackage{amsmath} 
\usepackage{booktabs} 
\usepackage{amssymb}
\usepackage{times}  
\usepackage{helvet}  
\usepackage{courier}  
\usepackage[hyphens]{url}  
\usepackage{graphicx} 
\usepackage{natbib}  
\usepackage{caption} 
\usepackage{algorithm}
\usepackage{algorithmic}
\usepackage{xcolor, multirow, colortbl}

\newcommand{\best}[1]{\cellcolor{red!10}{#1}}
\newcommand{\second}[1]{\cellcolor{blue!10}{#1}}
\definecolor{mygrey}{cmyk}{0, 0, 0, 0.5}
\newcommand{\smallgrey}[1]{\textcolor{mygrey}}

\usepackage{newfloat}
\usepackage{listings}
\DeclareCaptionStyle{ruled}{labelfont=normalfont,labelsep=colon,strut=off} 
\floatstyle{ruled}
\newfloat{listing}{tb}{lst}{}
\floatname{listing}{Listing}
\nocopyright 

\title{CAT: A Conditional Adaptation Tailor for Efficient and Effective Instance-Specific Pansharpening on Real-World Data}
\author{
    Tianyu Xin\equalcontrib,
    Jin-Liang Xiao\equalcontrib,
    Zeyu Xia,
    Shan Yin,
    Liang-Jian Deng\thanks{Corresponding author.}
}
\affiliations{
    \textsuperscript{\rm 1}University of Electronic Science and Technology of China\\

    tyxin@std.uestc.edu.cn, jinliang\_xiao@163.com, liangjian.deng@uestc.edu.cn
}

\usepackage{bibentry}

\begin{document}

\maketitle

\begin{abstract}
Pansharpening is a crucial remote sensing technique that fuses low-resolution multispectral (LRMS) images with high-resolution panchromatic (PAN) images to generate high-resolution multispectral (HRMS) imagery. Although deep learning techniques have significantly advanced pansharpening, many existing methods suffer from limited cross-sensor generalization and high computational overhead, restricting their real-time applications. 
  To address these challenges, we propose an efficient framework that quickly adapts to a specific input instance, completing both training and inference in a short time. Our framework splits the input image into multiple patches, selects a subset for unsupervised CAT training, and then performs inference on all patches, stitching them into the final output. The CAT module, integrated between the feature extraction and channel transformation stages of a pre-trained network, tailors the fused features and fixes the parameters for efficient inference, generating improved results.
  Our approach offers two key advantages: (1) \textit{Improved Generalization Ability}: by mitigating cross-sensor degradation, our model—although pre-trained on a specific dataset—achieves superior performance on datasets captured by other sensors; (2) \textit{Enhanced Computational Efficiency}: the CAT-enhanced network can swiftly adapt to the test sample using the single LRMS-PAN pair input, without requiring extensive large-scale data retraining. Experiments on the real-world data from WorldView-3 and WorldView-2 datasets demonstrate that our method achieves state-of-the-art performance on cross-sensor real-world data, while achieving both training and inference of $512\times512$ image within \textit{0.4 seconds} and $4000\times4000$ image within \textit{3 seconds} at the fastest setting on a commonly used RTX 3090 GPU.
\end{abstract}


\section{Introduction} \label{sec:intro}
Remote sensing images are extensively utilized in various domains, such as environmental monitoring \cite{env1, env2}, urban planning \cite{urban1, urban2}, and precision agriculture \cite{agriculture1,agriculture2}. However, due to hardware limitations, satellite sensors cannot capture images that are both spectrally rich and of high spatial resolution. Typically, satellites like WorldView series capture low-resolution multispectral (LRMS) images and high-resolution panchromatic (PAN) images separately. Consequently, pansharpening aims to fuse an LRMS image with a high-resolution PAN image to generate a high-resolution multispectral (HRMS) image that preserves both spectral and spatial information, which plays a vital role in the field of remote sensing image processing.

\begin{figure}[t]
  \centering
  \includegraphics[width=0.5\textwidth]{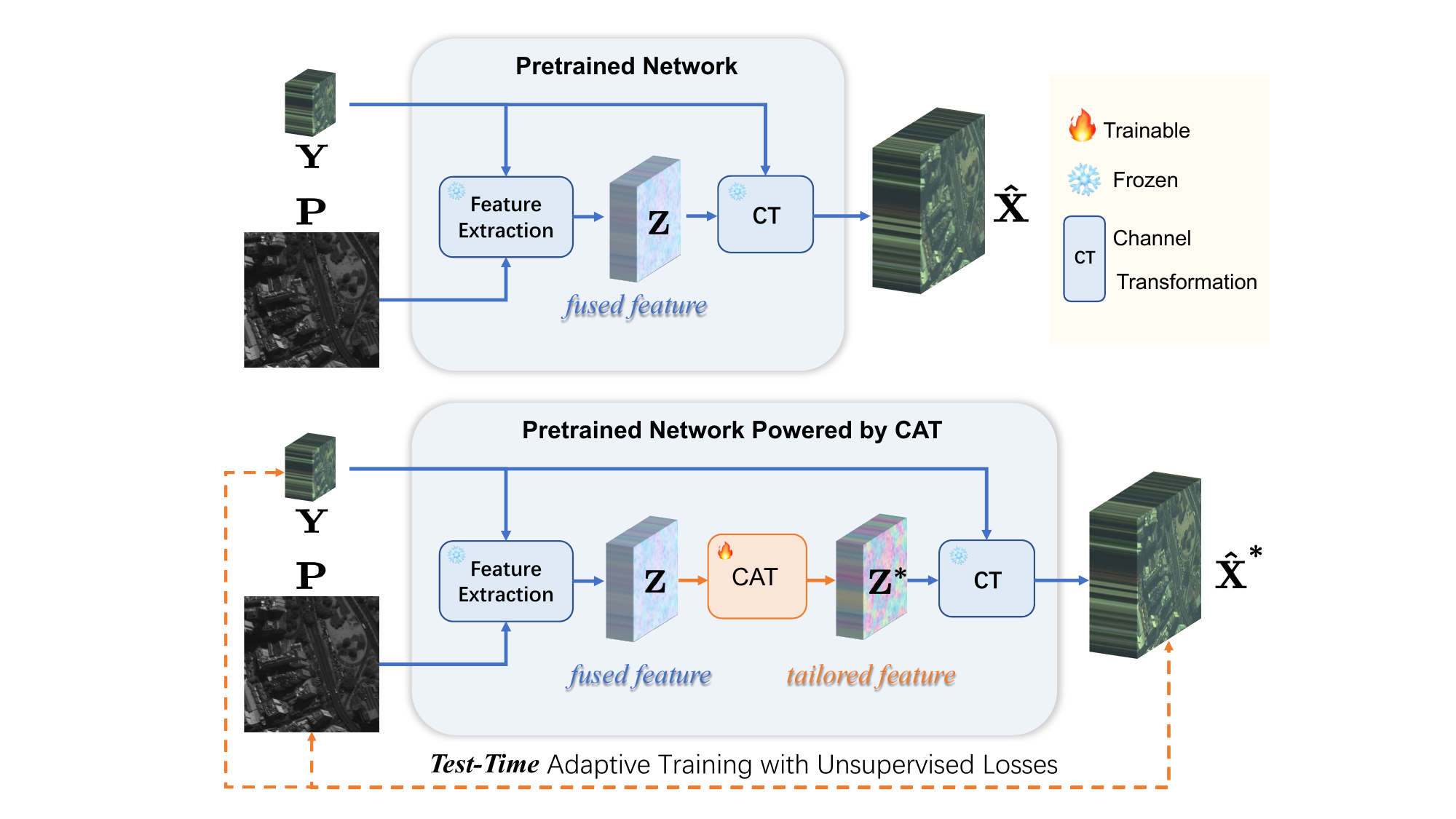}
  \caption{CAT-enhanced Network: CAT module is integrated into a pre-trained network between feature extraction layers and channel transformation layers, tailoring the high-dimensional fused feature. This design enables swift adaptation training using only the single input LRMS-PAN pair ($\mathbf Y$-$\mathbf P$) and generates improved HRMS results ($\mathbf{\hat X}^*$). }
  \label{fig:method_overview}
\end{figure}

Pansharpening methods are broadly classified into four categories: component substitution (CS), multi-resolution analysis (MRA), variational optimization (VO), and deep learning (DL) approaches \cite{deng2022machine, Vivone24, vivone2020new}. The first three categories are traditional approaches that require only a PAN-LRMS image pair as input. Although CS and MRA methods are computationally efficient, they often suffer from spectral and spatial distortions, respectively. VO approaches rely on the formulation of a variational model. However, they are generally limited in capturing the nonlinear relationships between PAN and HRMS images, which constrains their overall fusion performance \cite{xiao2022, xiao2023variational}. In contrast, deep learning methods leverage neural network architectures excellent feature extraction capability, such as convolutional neural network (CNN)-based models (e.g., \cite{dicnn, pannet, cao2021pancsc, yang2023panflownet, wang2024deep}), Transformer-based models (e.g., \cite{panformer,fusformer, zhang2021gtp, wu2025fully}), and Diffusion-based methods (e.g., \cite{meng2023pandiff, cao2024diffusion, zhong2024ssdiff, rui2024unsupervised}), enabling them to produce superior fusion results. 

Despite their promising results, current DL methods encounter two major challenges: $1)$ \textit{limited generalization across different sensors and imaging scenarios}; $2)$ \textit{a data-dependent and time-consuming training process.} Specifically, DL models are typically trained on pre-collected datasets. When test samples exhibit distributional shifts relative to the training data, the performance of these models tends to degrade significantly. especially for real-world data, the gap between real-world and simulated data exacerbates the generalization deficiency of DL methods. Moreover, the effectiveness of DL approaches heavily depends on the availability of large-scale annotated datasets—resources that are often scarce on real-world remote sensing data. The substantial computational cost and time required for training further limit the practicality and scalability of deep learning-based methods in operational settings.

To address these challenges, we propose an innovative pansharpening framework shown in Figure \ref{fig:method_overview} that incorporates a lightweight, test-time adaptive module, termed the \textbf{Conditional Adaptation Tailor (CAT)}. Unlike conventional deep learning approaches that require either time-consuming retraining or computationally intensive zero-shot inference \cite{psdip}, our CAT module is integrated and updated exclusively at test time based solely on the input data, eliminating the need for large-scale retraining. This design seamlessly integrates the CAT module into various pre-trained backbone networks without altering their original parameters, preserving the pretrained model’s performance while enabling rapid adaptation to new data. Specifically, the CAT module is trained in a physical-information-motivated unsupervised manner and is positioned between the backbone’s feature extraction and channel transformation stages, where it learns residual adjustments in the high-dimensional latent representations generated by the early layers. This unsupervised training approach, combined with its strategic placement, allows CAT to efficiently tailor intermediate features to match domain shifts encountered at test time, significantly enhancing cross-dataset and cross-sensor generalization while retaining valuable pretrained knowledge. Furthermore, we employ a patch-wise training and inference strategy: a small number of randomly selected image patches are used to efficiently optimize the CAT parameters, and parallel inference across all patches minimizes overall processing time, enabling high-quality pansharpening of large-scale, megapixel satellite imagery within seconds. With unsupervised learning and conditional feature adaptation, our framework delivers significant gains in both performance and computational efficiency, offering a practical and scalable solution for real-time remote sensing applications. Experiments demonstrate that the proposed method with CAT achieves state-of-the-art performance in real-time application on real-world data.

The main contributions are summarized as follows:
\begin{itemize}
  \item We propose a novel real-world-based pansharpening framework that can integrates a lightweight, test-time adaptive module, i.e., the Conditional Adaptation Tailor (CAT), into pretrained networks.
  \item Our approach dramatically accelerates adaptive training by optimizing only the lightweight CAT module on a limited set of selected patches, combined with parallel inference—substantially reducing computational cost and complexity.
  \item Extensive experiments demonstrate that CAT-enhanced models achieve SOTA results on both pretrained dataset and other dataset within a second per image, proving both efficiency and efficacy. SOTA results on megapixel image within 3 seconds also indicates its furture potential. 
\end{itemize}

\section{Related Works}
\subsection{Deep Learning-based Pansharpening}

Deep learning has substantially enhanced pansharpening performance by enabling models to capture complex spatial-spectral relationships through various neural network architectures. CNN-based methods such as PanNet \cite{pannet}, DiCNN \cite{dicnn}, and FusionNet \cite{fusionnet} have shown strong capabilities in enhancing spatial details while preserving spectral fidelity. Building on this foundation, Transformer-based approaches \cite{panformer, fusformer} and diffusion-based models \cite{pandiff, ssdiff} further improve performance by leveraging more sophisticated architectures to extract and fuse spatial and spectral features effectively.

\textit{Despite their impressive results, most deep learning-based methods suffer from limited generalization.} While these models perform well when the test data originates from the same sensor as the training data, their effectiveness degrades significantly when applied to images acquired from different sensors or satellites. This cross-domain performance gap presents a critical challenge for practical deployment in diverse scenarios.

\subsection{Instance-Specific Pansharpening}

Instance-specific pansharpening methods generate high-resolution multispectral (HRMS) images solely from a single LRMS-PAN input pair—using that same pair for both training and inference—without relying on extensive large-scale datasets. These methods can be broadly classified into two categories: traditional pansharpening techniques and zero-shot pansharpening approaches.

Traditional pansharpening methods, including CS \cite{cs1, cs2}, MRA \cite{mra1, mra2}, and VO methods \cite{vo1, vo2}, only require a single PAN-LRMS pair to generate the fused HRMS image. \textit{Efficient as they are, their pansharpening capabilities are often inferior than deep learning-based methods} since traditional approaches rely heavily on handcrafted transformations and struggle to capture the complex, nonlinear spatial-spectral relationships inherent in remote sensing data. 

On the other hand, zero-shot pansharpening approaches also only require PAN-LRMS pair as input, but also adequately facilitate superior pansharpening capabilities of neural networks. Zero-shot pansharpening frameworks like PSDip \cite{psdip} and ZS-Pan \cite{zspan} generally exhibit improved results over traditional approaches in terms of instance-specific pansharpening. \textit{Despite efficacy, their processing time towards a single PAN-LRMS pair always prolongs to minute-level, which means their training time is hundreds time longer than inference, making it hardly practical for real-time application. }

\section{Methodology}

To overcome the challenges of limited generalization and high computational overhead on new test data, we introduce the CAT module to efficiently enhance various pre-trained networks. Leveraging an unsupervised training approach combined with its feature tailoring capability, the CAT-enhanced network consistently delivers improved performance. Moreover, by employing patch-wise adaptation and parallel inference strategies, our model significantly reduces both computational overhead and processing time, while achieving performance improvements. 

\subsection{Overall Framework} \label{sec:overall_framework}

\begin{figure*}[t]
  \centering
  \includegraphics[width=0.95\textwidth]{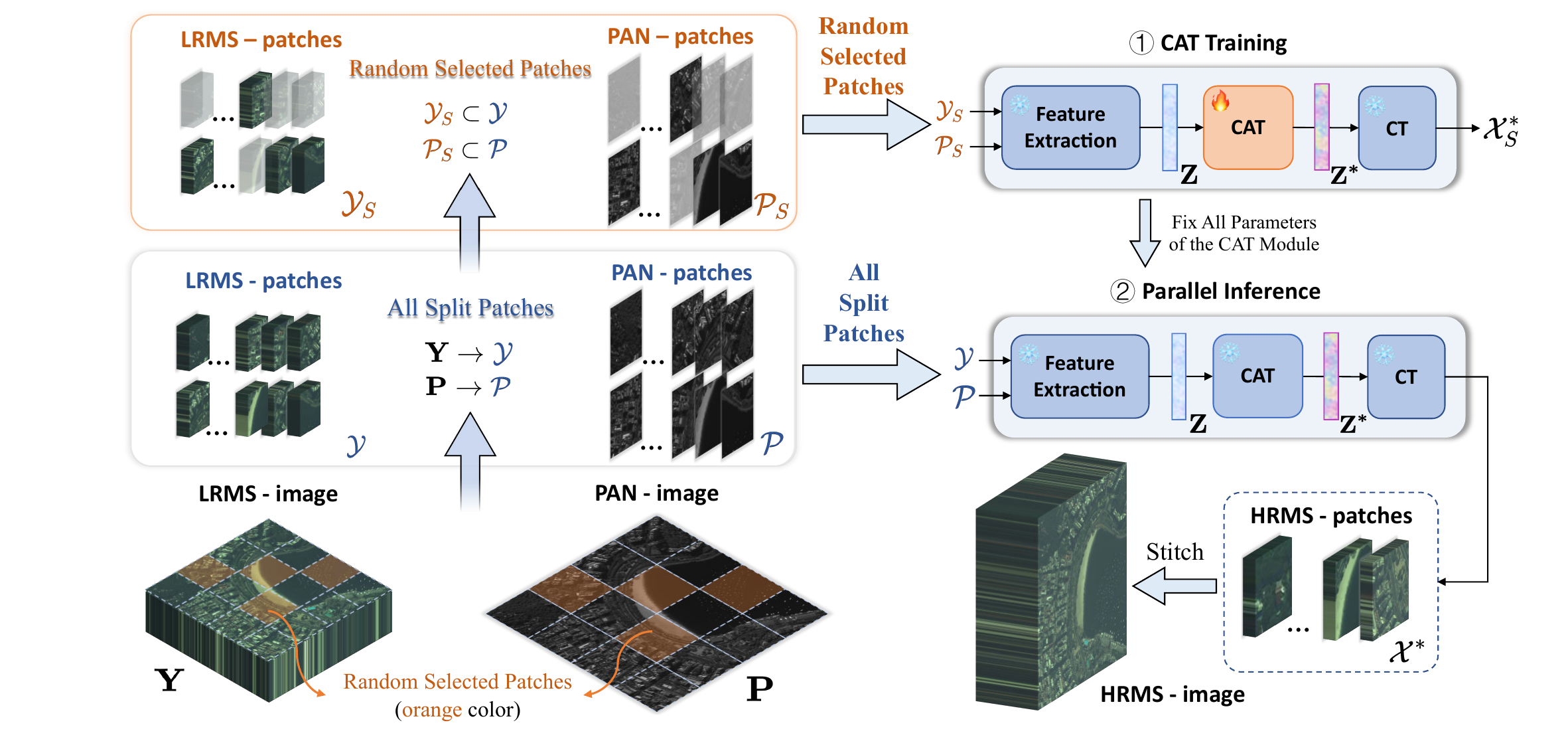}
  \caption{Overall Framework of the CAT-enhanced Network: The input LRMS-PAN pair is partitioned into multiple patches, where a random subset is selected for unsupervised CAT training. With the CAT module parameters fixed, the network then conducts parallel inference on all patches, and the resulting HRMS patches are stitched together to form the final HRMS image. }
  \label{overall_framework}
\end{figure*}

The overall framework of our proposed method is illustrated in Figure \ref{overall_framework}. With the original PAN and LRMS image split to multiple patches, the process firstly conducts CAT training on random selected patches with an unsupervised approach. Once CAT module finishes the adaptive training, parallel inference is performed on all split patches, which are subsequently stitched together to produce the final fused HRMS image.

Given the original low-resolution multispectral (LRMS) image and panchromatic (PAN) image, they are first split into $N$ non-overlapping patches. The LRMS patches and PAN patches are spatially paired, with the sets of LRMS patches and PAN patches denoted as $\mathcal{Y}$ and $\mathcal{P}$, respectively. Then from these patches, a random subset $\mathcal Y_S \subset \mathcal Y$ and $\mathcal P_S \subset \mathcal P$ is selected for CAT module training. 

The CAT (Conditional Adaptation Tailor) module is introduced at a strategically chosen position inside the pre-trained backbone network, allowing it to tailor the high-dimensional representations generated by backbone network. By making these adjustments, the CAT module enhances the final output, producing a high-resolution multispectral (HRMS) image with improved quality and accuracy. After training the CAT module, the network performs parallel inference on all LRMS-PAN patches with fixed parameters. All output patches are then stitched together to form the final HRMS image.

\subsection{CAT Module Position and Structure Design} \label{sec:position_and_structure}

In this section, we provide a detailed explanation for the placement and structural design of the CAT (Conditional Adaptation Tailor) module within our pansharpening framework.

\subsubsection{Placement Design}

For the pansharpening task, DL methods can be generally described by the following equation: 
\begin{equation}
  \mathbf{\hat X} = \mathcal F(\mathbf Y, \mathbf P; \theta),
  \label{ori_infer}
\end{equation}
where $\mathbf{\hat X} \in \mathbb{R}^{C \times H \times W}$ denotes the fused HRMS inferred by the original backbone network $\mathcal F$, and $\theta$ denotes the parameters of backbone network. Despite the diverse designs of the backbone network $\mathcal{F}$ across different approaches, many methods (e.g., \cite{pannet,dicnn,fusionnet,pmac}) adopt a similar structure that can be easily decomposed into a feature extraction stage and a channel transformation stage. The feature extraction stage is typically implemented by the early layers of the network, denoted as $\mathcal{F}_1$ with parameters $\theta_1$, while the channel transformation stage is handled by the later layers, denoted as $\mathcal{F}_2$ with parameters $\theta_2$. Consequently, the overall network can be expressed as $\mathcal{F} = \mathcal{F}_2 \cdot \mathcal{F}_1$, with the complete set of parameters given by $\theta = \theta_1 \cup \theta_2$.

In the feature extraction stage, early layers of the backbone network extracts spatial and spectral features from the PAN and LRMS inputs and fuses them into a high-dimensional latent representation. The strategy for feature extraction may vary across different networks, but the goal remains the same. Mathematically, this stage can be expressed as:
\begin{equation}
  \mathbf Z = \mathcal F_1(\mathbf Y, \mathbf P; \theta_1),
  \label{infer1}
\end{equation}
where $\mathbf Z \in \mathbb{R}^{S \times H \times W}$ is the high-dimension latent representation with $S$ channels. 

The channel transformation stage follows, where the fused features $\mathbf Z$ are transformed into the final HRMS image by later layers of the backbone network. This is typically achieved by projecting the high-dimensional latent representation back to the target spectral space. The transformation is modeled as:
\begin{equation}
  \mathbf{\hat X} = \mathcal F_2(\mathbf Z; \theta_2) + \texttt{UpSample}(\mathbf Y),
  \label{channel_transformation}
\end{equation}
where \texttt{UpSample}$(\mathbf Y)$  denotes the upsampled LRMS image used as a residual connection to enhance spectral consistency.

\begin{figure}[ht]
    \centering
    \includegraphics[width=0.9\linewidth]{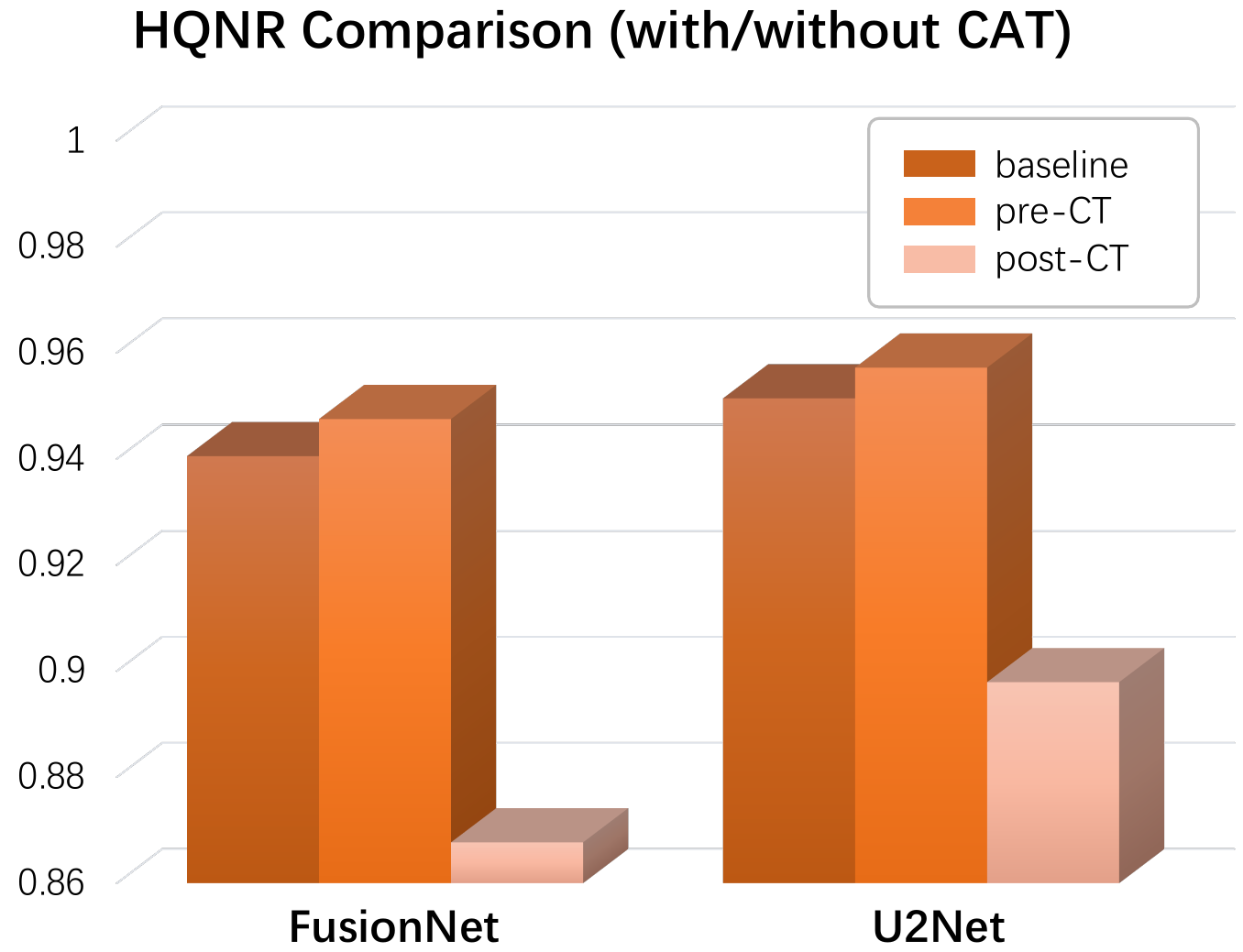}
    \caption{HQNR comparison for FusionNet and U2Net under three CAT placement strategies: CAT not inserted (Baseline), CAT inserted before channel transformation (pre-CT), and CAT inserted after channel transformation (post-CT). The pre-CT configuration consistently outperforms the others, validating our chosen CAT placement.}
    \label{fig:position_cmp}
\end{figure}

To enhance cross-sensor and cross-dataset generalization, the CAT module is inserted between the feature extraction and channel transformation stages, which also ensures our strategy compatible with a wide range of backbone network. Positioned at this junction, the CAT module operates on the rich, high-dimensional latent representation $\mathbf{Z}$, learning residual adjustments that tailor the features to match the test-time data distribution. The refined latent representation is given by:
\begin{equation}
  \mathbf{Z}^* = \mathcal{G}(\mathbf Z; \phi) + \mathbf Z,
  \label{Z_tailoring}
\end{equation}
where $\mathcal{G}$ denotes the residual learning part of CAT module with trainable parameters $\phi$, and $\mathbf{Z}^*$ denotes the tailored representation. 

This strategic placement enables the CAT module to effectively refine the intermediate features, thereby improving the final HRMS output:
\begin{equation}
  \mathbf{\hat X^*} = \mathcal F_2(\mathbf Z^*; \theta_2) + \texttt{UpSample}(\mathbf Y),
\end{equation}
where $\mathbf{\hat X^*}$ denotes the adapted HRMS image produced using the tailored representation $\mathbf Z^*$. 

To further validate the rationality of our CAT module placement, we conducted experiments comparing model performance when the CAT module is integrated at different positions within the pre-trained network. Given the variability in feature extraction architectures and our intent for broad applicability, we deliberately avoid injecting the CAT module into the feature extraction layers. Instead, we examine two placement options: inserting the CAT module before channel transformation (pre-CT, which is our current approach) and after channel transformation (post-CT). As shown in Figure \ref{fig:position_cmp}, the pre-CT placement consistently achieves higher HQNR values compared to both the baseline and the post-CT configuration. These results confirm that positioning the CAT module prior to the channel transformation stage enables it to effectively tailor the high-dimensional fused features, thereby enhancing overall performance and affirming the generalizability of our method.

\subsubsection{Structure Design} 

\begin{figure*}[t]
  \centering
  \includegraphics[width=0.9\textwidth]{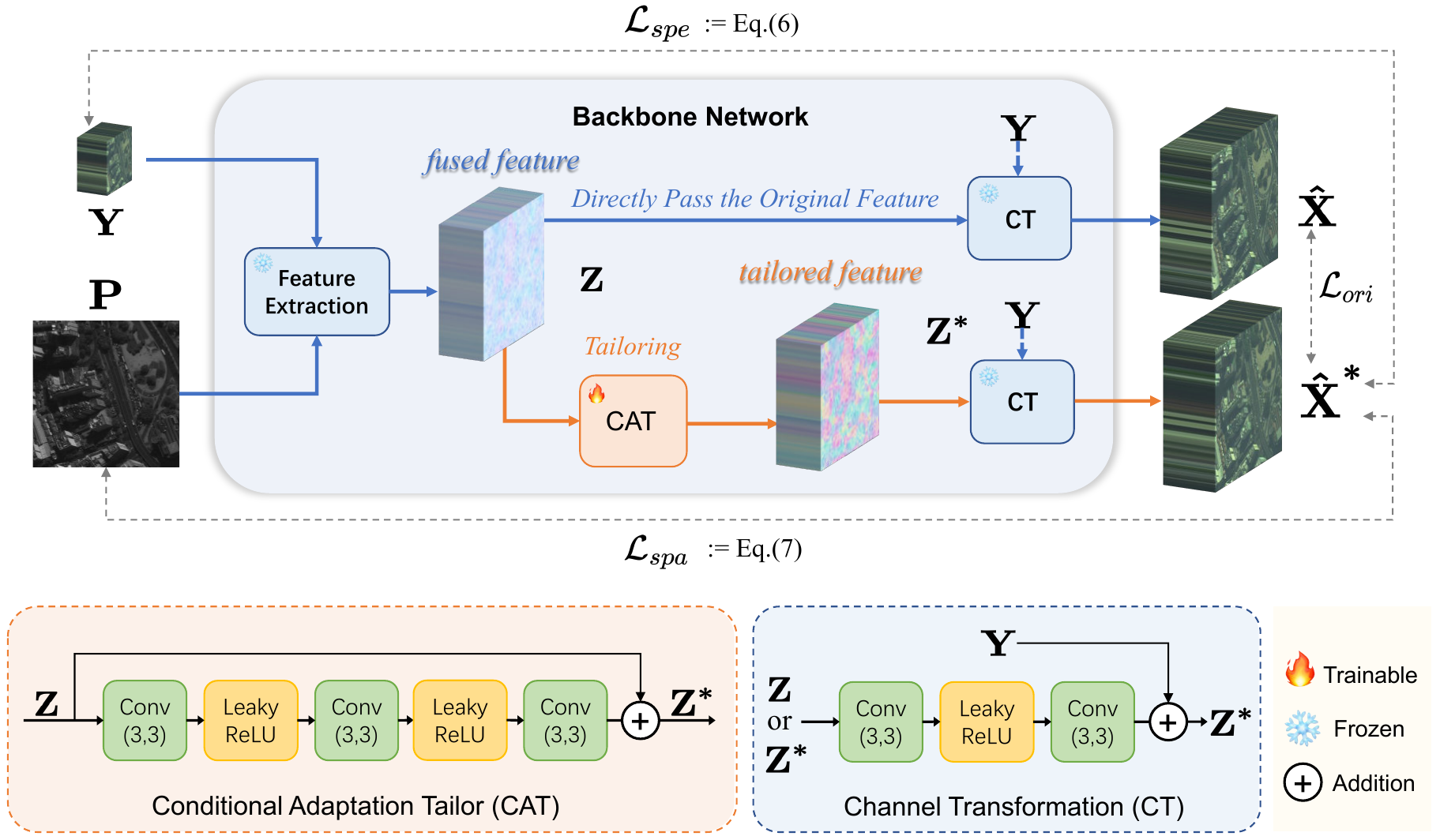}
  \caption{CAT Module Training : LRMS image $\mathbf Y$ and PAN image $\mathbf P$ are passed into the backbone network, where they are transformed to high-dimensional representations in latent space, denoted as $\mathbf Z$ and then tailored by CAT module, generating tailored representation $\mathbf Z^*$. Both the original and tailored representations are passed through CT module to the channel space, generating original HRMS $\mathbf{\hat X}$ and tailored HRMS $\mathbf{\hat X}^*$. }
  \label{fig:network_details}
\end{figure*}

The internal architecture of the CAT module is implemented as a few-layer convolutional network with a residual connection, as illustrated in Figure \ref{fig:network_details}. This compact design is motivated by functional similarity and continuity with the existing network, enabling the CAT module to process high-dimensional latent representations and learn residual corrections that compensate for deviations induced by domain shifts during test time (see Eq.~(\ref{Z_tailoring})). This adaptive adjustment is conceptually analogous to the channel transformation performed by $\mathcal{F}_2$, which maps latent features into the desired spectral domain while incorporating a residual addition.

A key advantage of our approach lies in its efficiency. With a small parameter footprint and a simple few-layer structure, the CAT module enables rapid updates, allowing for real-time adaptation that is significantly faster than more complex modules based on, for example, attention mechanisms.

Structurally, designing the CAT module with an architecture similar to that of $\mathcal{F}_2$ ensures seamless integration into the pre-trained network. This architectural continuity not only simplifies the integration process but also promotes smooth information flow between the feature extraction and channel transformation stages, ultimately enhancing the overall efficiency and adaptability of the network.

\subsection{Unsupervised CAT Training Loss} \label{sec:updating_method}

To efficiently adapt pre-trained networks to new datasets without labels, we propose an unsupervised CAT training approach guided by three carefully designed loss functions. Specifically, we introduce a spectral loss ($\mathcal{L}_{spe}$), a spatial loss ($\mathcal{L}_{spa}$), and an original-output consistency loss ($\mathcal{L}_{ori}$). The spectral and spatial losses aim to enforce meaningful correspondence with input data, while the consistency loss regularizes the adaptation, preventing overfitting and excessive deviation from the pre-trained model's original output. These loss functions collectively ensure spectral consistency, spatial detail preservation, and stability of the training process. 

\subsubsection{Spectral Loss ($\mathcal L_{spe}$)} 

The spectral loss is introduced to preserve spectral consistency between the generated high-resolution multispectral (HRMS) output and the corresponding low-resolution multispectral (LRMS) input. To achieve this, we spatially blur the tailored HRMS output to the resolution of the input LRMS image and calculate the spectral discrepancy as:
\begin{equation}
  \label{spectral_loss}
  \mathcal{L}_{spe} = \|\texttt{DownSample}(\mathbf{\hat X^*} \mathbf B) - \mathbf{Y}\|_1,
\end{equation}
where $\mathbf{\hat X^*}$ denotes the tailored HRMS generated by the model, $\mathbf B$ denotes the blurring operation achieved by modulation transfer functions (MTFs), $\texttt{DownSample}$ denotes the downsampling process, and $\|\cdot\|_1$ means $\ell_{1}$-norm.

\subsubsection{Spatial Loss ($\mathcal{L}_{spa}$)}

The spatial loss encourages the tailored HRMS image to retain high-frequency spatial details consistent with structural information provided by the PAN image. Similar to the methodology presented in \cite{vo+net}, the spatial loss is calculated by extracting high-frequency components from both the PAN image and the luminance channel of the generated HRMS image and measuring their difference:

\begin{equation}
  \label{spatial_loss}
  \mathcal{L}_{spa} = \|\mathbf{\hat X^*} - \mathbf{\hat X^*} \mathbf B \circ (\mathbf{\hat P} \oslash \mathbf{\hat P} \mathbf B)\|_1,
\end{equation}
where $\mathbf{\hat P} \in \mathbb R^{C \times H \times W}$ denotes reference panchromatic image acquired by PAN image extended to each channel $C$, $\circ$ denotes element-wise multiplication, and $\oslash$ represents element-wise division. 

\subsubsection{Original-Output Consistency Loss ($\mathcal{L}_{ori}$)}

To prevent excessive deviation and potential overfitting during adaptation, we introduce a consistency loss to regularize the CAT-enhanced output, ensuring it remains close to the pre-trained backbone's original output:
\begin{equation}
  \label{original_loss}
  \mathcal{L}_{ori} = \|\mathbf{\hat X^*} - \mathbf{\hat X}\|_1,
\end{equation}
where $\mathbf{\hat X^*}$ denotes tailored HRMS image generated by CAT-enhanced network, and $\mathbf{\hat X}$ denotes HRMS directly generated by backbone network without being tailoring. 

\subsubsection{Overall Loss}

Finally, parameters of the CAT module, denoted as $\phi$, are optimized by minimizing a weighted combination of the three aforementioned losses:

\begin{equation}
  \mathcal{L}_{total} = \eta_{1}\mathcal{L}_{spe} + \eta_{2}\mathcal{L}_{spa} + \eta_{3}\mathcal{L}_{ori},  
\end{equation}
where $\eta_{1}$, $\eta_{2}$, and $\eta_{3}$ are hyperparameters controlling the relative importance of each loss component.

This unsupervised loss formulation ensures rapid yet effective training, allowing the CAT module to achieve superior performance and excellent generalization across diverse scenarios without incurring significant computational overhead.

\subsection{Patch-wise Adaptation and Parallel Inference} \label{sec:patch_and_parallel}

To significantly improve computational efficiency and scalability, particularly for high-resolution satellite imagery, we propose a patch-wise adaptation and parallel inference strategy. Instead of processing entire images, each PAN-LRMS input pair is partitioned into multiple patches. To achieve adaptation, only a selected subset of these patches is used for updating parameters of the CAT module through unsupervised loss functions. Once the update completes, its parameters are fixed and applied to perform inference on all patches simultaneously in parallel. The combination of patch-wise adaptation and parallel inference minimizes computational overhead during adaptation and leverages parallel execution for rapid, scalable processing.

Formally, given a PAN-LRMS pair with a resolution of $H \times W$ for the panchromatic image and $(H/r) \times (W/r)$ for the low-resolution multispectral (LRMS) image, we split the images into $N$ patches:
\begin{equation}
  \text{PatchPartition}(\mathbf Y) = \mathcal Y = \{\mathbf y_i | i = 1, 2, \dots, N\},
\end{equation}
\begin{equation}
  \text{PatchPartition}(\mathbf P) = \mathcal P = \{\mathbf p_i | i = 1, 2, \dots, N\},
\end{equation}
where patch-LRMS $y_i \in \mathbb R^{C \times h \times w}$ and patch-PAN $p_i \in \mathbb R^{h \times w}$. 

During the adaptation stage, we randomly select a subset of $n$ patches from the total $N$ patches. Let the subset of selected patches denoted as $\mathcal S$. 
\begin{equation}
  \mathcal Y_S = \text{RandomSelect}(\mathcal Y) = \{\mathbf y_j|\mathbf y_j \in \mathcal Y, j = 1, 2, \dots, n\},
\end{equation}
\begin{equation}
  \mathcal P_S = \text{RandomSelect}(\mathcal P) = \{\mathbf p_j|\mathbf p_j \in \mathcal P, j = 1, 2, \dots, n\}.
\end{equation}
These patches serve as a representative subset for adaptation. By optimizing the parameters of the CAT module exclusively on the patches in $\mathcal{S}$ with unsupervised approaches, we efficiently update the CAT module without incurring excessive computational overhead. This selective, patch-based updating reduces the computational complexity from $O(N)$ to $O(n)$, thereby significantly accelerating the adaptation process.

After the CAT module has been updated using the selected patches, we perform inference on all $N$ patches independently and in parallel. Let $\mathbf{p}_i$ denotes the $i$-th PAN-patch and $\mathbf y_i$ denotes the $i$-th LRMS-patch from the original PAN-LRMS pair, and let $\mathcal{F}^*$ represent the CAT-enhanced network. The inference on each patch is then given by:
\begin{equation}
  \mathbf{\hat x^*}_i = \mathcal{F}^*(\mathbf{y}_i, \mathbf{p}_i), i = 1, 2, \ldots, N.
\end{equation}
The final high-resolution multispectral (HRMS) image is obtained by stitching together the inferred patches according to their original spatial arrangement:
\begin{equation}
  \mathbf {\hat X^*} = \texttt{Stitch}(\mathcal X^*), \mathcal X^* = \{\mathbf{\hat x^*_i}|\mathbf{\hat x^*}_i, i=1, 2, \dots, N\},
\end{equation}
where $\mathcal X^*$ denotes the set of all inferred patch-HRMS, $\mathbf {\hat X^*}$ denotes the final stitched HRMS, and \texttt{Stitch} represents the spatially-aware stitching process. 

This parallelized inference approach significantly speeds up processing by leveraging the multi-core computing capabilities of modern devices, ensuring rapid and scalable performance on large-scale, high-resolution satellite imagery.

\section{Experiments}
 
\subsection{Datasets, Metrics, and Training Details}

\paragraph{\textbf{Datasets:}}
Our experiments employ two datasets derived from PanCollection\footnote{\url{https://github.com/liangjiandeng/PanCollection}}, captured by WorldView-3 (WV-3) and WorldView-2 (WV-2), and constructed in accordance with Wald’s protocol \cite{wald}. These datasets consist of panchromatic (PAN) images and low-resolution multispectral (LRMS) images, with PAN images sized at $512 \times 512$. Notably, our method specifically targets real-world pansharpening challenges—data of paramount practical significance and precisely the type for which our unsupervised loss function was designed. Consequently, we primarily present results on real-world data in the following sections. Additionally, we extend our evaluation to a megapixel image (PAN size: $4000 \times 4000$).

\paragraph{\textbf{Metrics:}}
We use well-established evaluation metrics to assess our method quantitatively. For real-world data, we employ HQNR \cite{aiazzi2014full}, $D_s$ and $D_{\lambda}$, where HQNR is derived from $D_s$ and $D_{\lambda}$ to provide a comprehensive assessment of image quality. 

\paragraph{\textbf{Training Details:}}
Our experimental framework employs FusionNet \cite{fusionnet} (2020) and U2Net \cite{u2net} (2023) pretrained on WV3 dataset as our baseline backbone networks, representing typical CNN-based and attention-based architectures, respectively. This selection underscores that our method is versatile and can be seamlessly integrated across diverse network designs, enhancing pansharpening performance regardless of the architecture used in the feature extraction layers.
And experiments were conducted on a hardware setup comprising an NVIDIA RTX 3090 GPU with 24GB memory and Intel i9-12900 CPU.
\textit{More details can be found in supplementary materials.} 

\begin{table*}[htbp]
  \centering 
  \setlength{\tabcolsep}{4pt}
  \renewcommand\arraystretch{1.2}
  \resizebox{0.95\linewidth}{!}{
    \begin{tabular}{c|ccc|ccc}
      \toprule
      \toprule
      \multirow{2.2}{*}{Method} & \multicolumn{3}{c|}{WV-3 (Real-World Data): Avg$\pm$std} & \multicolumn{3}{c}{WV-2 (Real-World Data): Avg$\pm$std} \\
      \cmidrule{2-4}\cmidrule{5-7}
       & HQNR$\uparrow$ & $D_\lambda\downarrow$ & $D_s\downarrow$ & HQNR$\uparrow$ & $D_\lambda\downarrow$ & $D_s\downarrow$ \\
      \midrule
          BT-H $^{\small\textcolor{gray}{2017}}$        & 0.8659$\pm$0.0568 & 0.0656$\pm$0.0262 & 0.0742$\pm$0.0383 & 0.8300$\pm$0.0430 & 0.0860$\pm$0.0301 & 0.0925$\pm$0.0208 \\
          C-BDSD $^{\small\textcolor{gray}{2014}}$     & 0.8562$\pm$0.0233 & 0.0874$\pm$0.0236 & 0.0618$\pm$0.0128 & 0.6956$\pm$0.0461 & 0.2253$\pm$0.0488 & 0.1019$\pm$0.0296 \\
          BDSD-PC$^{\small\textcolor{gray}{2019}}$      & 0.8673$\pm$0.0543 & 0.0634$\pm$0.0246 & 0.0749$\pm$0.0359 & 0.8286$\pm$0.0432 & 0.1413$\pm$0.0320 & $\second{0.0356\pm0.0213}$ \\
          MTF-GLP $^{\small\textcolor{gray}{2006}}$      & 0.9026$\pm$0.0444 & 0.0373$\pm$0.0124 & 0.0628$\pm$0.0359 & 0.8549$\pm$0.0475 & 0.0582$\pm$0.0221 & 0.0930$\pm$0.0320 \\
          MTF-GLP-FS $^{\small\textcolor{gray}{2018}}$ & 0.9127$\pm$0.0348 & 0.0357$\pm$0.0106 & 0.0537$\pm$0.0273 & 0.8658$\pm$0.0415 & 0.0563$\pm$0.0212 & 0.0830$\pm$0.0260 \\
          MF $^{\small\textcolor{gray}{2016}}$               & 0.9014$\pm$0.0358 & 0.0452$\pm$0.0121 & 0.0561$\pm$0.0268 & 0.8508$\pm$0.0538 & 0.0704$\pm$0.0308 & 0.0857$\pm$0.0297 \\
          PsDip $^{\small\textcolor{gray}{2024}}$         & 0.9215$\pm$0.0176 & 0.0191$\pm$0.0078 & 0.0607$\pm$0.0117 & 0.8980$\pm$0.0226 & $\best{0.0385\pm0.0239}$ & 0.0659$\pm$0.0158 \\
          ZS-Pan $^{\small\textcolor{gray}{2024}}$      & 0.9449$\pm$0.0208 & 0.0254$\pm$0.0071 & 0.0306$\pm$0.0153 & $\second{0.9112\pm0.0336}$ & 0.0476$\pm$0.0270 & 0.0435$\pm$0.0130 \\
          FusionMamba $^{\small\textcolor{gray}{2024}}$ & $\second{0.9550\pm0.0110}$ & 0.0186$\pm$0.0078 & $\second{0.0269\pm0.0058}$ & 0.9064$\pm$0.0227 & 0.0527$\pm$0.0252 & 0.0431$\pm$0.0114 \\
          WFANet $^{\small\textcolor{gray}{2025}}$        & 0.9463$\pm$0.0095 & $\best{0.0162\pm0.0078}$ & 0.0381$\pm$0.0042 & 0.8922$\pm$0.0223 & 0.0633$\pm$0.0221 & 0.0476$\pm$0.0073 \\
          FusionNet $^{\small\textcolor{gray}{2020}}$  & 0.9405$\pm$0.0197 & 0.0229$\pm$0.0080 & 0.0375$\pm$0.0143 & 0.8881$\pm$0.0213 & 0.0543$\pm$0.0273 & 0.0606$\pm$0.0162 \\
          U2Net $^{\small\textcolor{gray}{2023}}$         & 0.9514$\pm$0.0114 & $\second{0.0178\pm0.0072}$ & 0.0313$\pm$0.0074 & 0.8706$\pm$0.0809 & 0.0936$\pm$0.0790 & 0.0400$\pm$0.0098 \\
          \midrule
          CAT$_{\text{FusionNet}}$     & 0.9475$\pm$0.0163 & 0.0190$\pm$0.0070 & 0.0342$\pm$0.0120 & 0.9099$\pm$0.0256 & $\second{0.0446\pm0.0214}$ & 0.0477$\pm$0.0160 \\
          CAT$_{\text{U2Net}}$        & $\best{0.9572\pm0.0132}$ & 0.0201$\pm$0.0073 & $\best{0.0232\pm0.0092}$ & $\best{0.9128\pm0.0526}$ & 0.0617$\pm$0.0494 & $\best{0.0274\pm0.0073}$ \\
      \bottomrule
      \bottomrule
    \end{tabular}
  }
  \caption{Performance comparison on real-world data of WorldView-3 (WV-3) and WorldView-2 (WV-2) datasets. The reported values represent the average performance over 20 test images. (\colorbox{red!10}{Red}: best; \colorbox{blue!10}{Blue}: second best)}
  \label{tab:wv3_wv2_results}
\end{table*}

\subsection{Benchmark}
To demonstrate the effectiveness of our proposed method, we compared it with several instance-specific pansharpening approaches, including BT-H \cite{BT-H}, C-BDSD \cite{C-BDSD}, BDSD-PC \cite{BDSD-PC}, MTF-GLP \cite{MTF-GLP}, MTF-GLP-FS \cite{MTF-GLP-FS}, and MF \cite{MF}. We also evaluated our method against state-of-the-art zero-shot pansharpening approaches, such as PsDip \cite{psdip} and ZS-Pan \cite{zspan}. Additionally, we included baseline models—namely, FusionNet \cite{fusionnet} and U2Net \cite{u2net}—to further validate the performance improvements achieved by our approach.

\subsection{ Main Experimental Results}
\begin{figure*}
    \centering
    \includegraphics[width=1\linewidth]{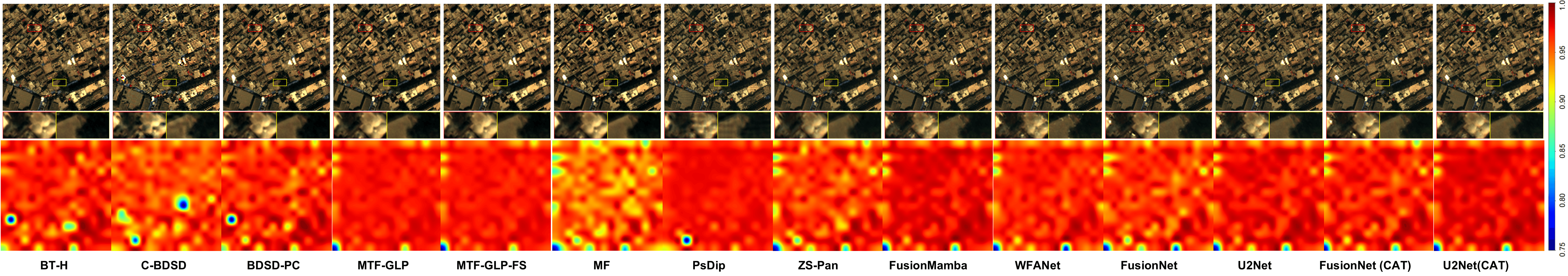}
    \caption{Visual Fusion Image and HQNR Map on WV-3 dataset (real-world data).}
    \label{fig:wv3_results}
\end{figure*}

\begin{figure*}
    \centering
    \includegraphics[width=1\linewidth]{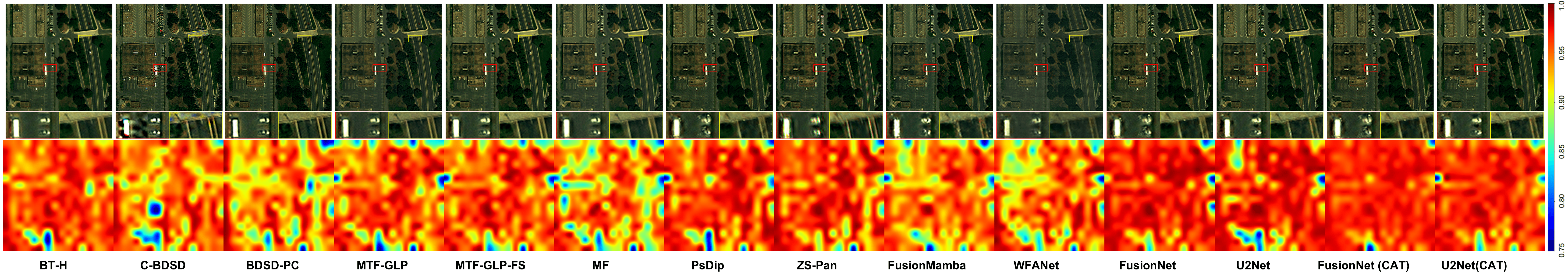}
    \caption{Visual Fusion Image and HQNR Map on WV-2 dataset (real-world data).}
    \label{fig:wv2_results}
\end{figure*}
To demonstrate the practical value of our method, we conducted comprehensive comparisons with a range of existing pansharpening techniques using real-world data from the WV-3 and WV-2 datasets. The results, summarized in Table \ref{tab:wv3_wv2_results}, present performance in terms of HQNR, $D_\lambda$, and $D_s$. The findings show that networks enhanced with the CAT module consistently outperform their baseline counterparts in terms of fusion quality. Moreover, the CAT-enhanced networks surpass state-of-the-art frameworks in HQNR, despite the baseline models exhibiting weaker fusion abilities.

To clearly illustrate the performance of each method, we present fusion result images and HQNR maps of the tested methods in Figures \ref{fig:wv3_results} and \ref{fig:wv2_results}, including a zoomed-in view of a specific region. Both the fusion images and HQNR maps demonstrate that the CAT-enhanced network consistently achieves the highest HQNR scores, further validating the effectiveness of the CAT module in improving pansharpening results across different datasets.

\subsection{Effectiveness and Generalization Ability}
\begin{figure}[h]
    \centering
    \includegraphics[width=0.85\linewidth]{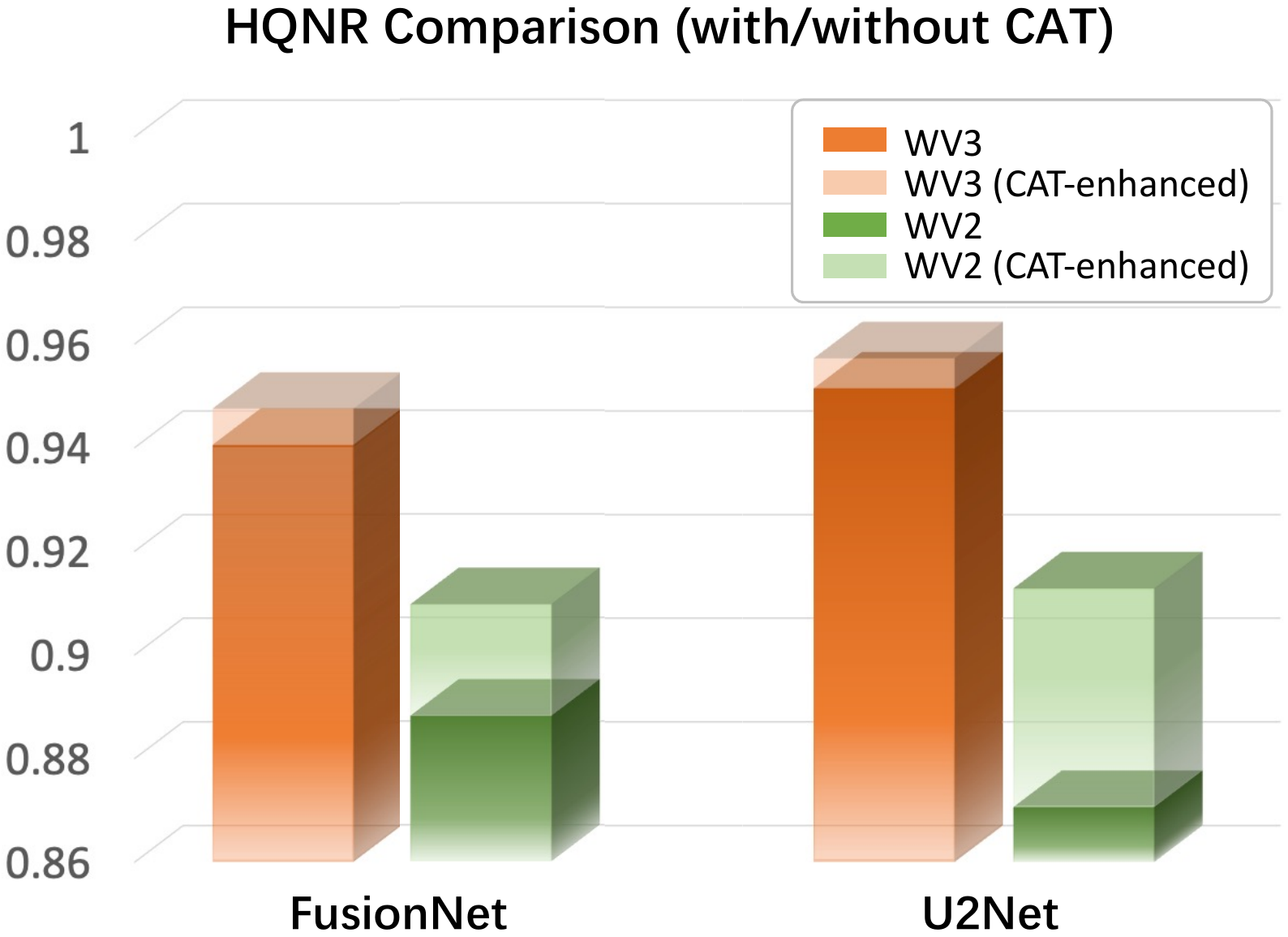}
    \caption{HQNR comparison of FusionNet and U2Net (both pretrained on the WV-3 dataset) with and without CAT enhancement on WV-3 and WV-2. The lighter-colored portions indicate the performance gain contributed by CAT.}
    \label{fig:ablation}
\end{figure}

\paragraph{\textbf{Effectiveness:}}
To validate the effectiveness of the proposed CAT-enhanced approach, we conduct a detailed comparison between baseline models and their CAT-enhanced counterparts. For the baseline model FusionNet, HQNR improvements of 0.0092 and 0.0218 were observed on the real-world data of WV3 and WV2, respectively. Similarly, for U2Net, HQNR gains of 0.0218 and 0.0422 were achieved on WV3 and WV2, respectively. These enhancements are clearly visualized in Figure \ref{fig:ablation}. Notably, as our framework introduces only a single CAT module into a pre-trained model, this comparison between configurations with and without CAT also serves as an effective ablation study, isolating the contribution of the CAT module to overall performance.

\paragraph{\textbf{Generalizability:}}
Conventional deep learning methods trained with large-scale supervision exhibit noticeable performance degradation when transitioning from the training set to a cross-sensor test set. As shown in Figure \ref{fig:ablation}, models trained on WV3 dataset exhibit reduced performance when applied to data of WV2. However, once enhanced with the CAT module (indicated by residuals with deeper color in Figure \ref{fig:ablation}), the performance of these deep learning methods improves substantially, underscoring the superior cross-sensor generalizability achieved by the CAT-enhanced networks.

\subsection{Discussions}
\paragraph{\textbf{Running Time:}}
While zero-shot methods generally outperform other instance-specific approaches and exhibit superior generalizability compared to deep learning models trained with large-scale supervision, they suffer from lengthy training and inference times, rendering real-time applications impractical. As shown in Table \ref{tab:time}, zero-shot pansharpening methods typically require over a minute to process a single LRMS-PAN image pair. In contrast, our CAT-enhanced network dramatically reduces processing time to less than one second while also achieving better performance. These results demonstrate that our method can deliver state-of-the-art results in a highly efficient and practical manner.
\begin{table}[!ht]
    \centering
    \setlength{\tabcolsep}{6pt}
    \renewcommand\arraystretch{1.2}
    \resizebox{\linewidth}{!}{
    \begin{tabular}{c|l|cccc}
         \toprule
         Dataset & Method & HQNR$\uparrow$ & \(D_\lambda\downarrow\) & \(D_s\downarrow\) & Duration$\downarrow$ (s) \\
         \midrule
         \multirow{4}{*}{WV-3} 
           
         & PsDip             & 0.9215 & \second{0.0191} & 0.0607 & 282.59    \\
         & ZS-Pan            & 0.9449 & 0.0254 & 0.0306 & 66.74    \\
         & $\text{CAT}_\text{FusionNet}$   & \second{0.9497} & \best{0.0183} & \second{0.0327} & \best{0.33}  \\
         & $\text{CAT}_\text{U2Net}$       & \best{0.9572} & 0.0201 & \best{0.0232} & \second{0.90}  \\
         \midrule
         \multirow{4}{*}{WV-2}
         & PsDip           & 0.8980 & 0.0385 & 0.0659 & 276.18    \\
         & ZS-Pan            & \second{0.9112} & \second{0.0476} & \second{0.0435} & 67.50    \\
         & $\text{CAT}_\text{FusionNet}$   & 0.9099 & \best{0.0446} & 0.0477 & \best{0.35}  \\
         & $\text{CAT}_\text{U2Net}$       & \best{0.9128} & 0.0617 & \best{0.0274} & \second{0.89}  \\
         \bottomrule
    \end{tabular}
    }
    \caption{Performance comparison of zero-shot methods and CAT-enhanced networks. (\colorbox{red!10}{Red}: best; \colorbox{blue!10}{Blue}: second best)}
    \label{tab:time}
\end{table}
\paragraph{\textbf{Results on Megapixel Images}}
Conventional deep learning methods are frequently constrained by lengthy training and inference times when processing large-scale data, rendering them impractical for real-time application of megapixel images (i.e., images comprising more than one million pixels). In contrast, our proposed approach enables deep learning methods to efficiently handle megapixel data without the need for extensive large-scale retraining.

To validate our method on megapixel-level data, we conducted experiments using a $4000\times4000$ PAN image and a $1000\times1000\times8$ LRMS image as inputs to generate a $4000\times4000\times8$ HRMS image, comprising a total of 128 million pixels. We then compared the results of our method with six classical instance-specific pansharpening methods.

\begin{table}[!ht]
    \centering
    \setlength{\tabcolsep}{6pt}
    \renewcommand\arraystretch{1.2}
    \resizebox{1\linewidth}{!}{
    \begin{tabular}{l|ccc|c}
         \toprule
         Method & HQNR$\uparrow$ & \(D_\lambda\downarrow\) & \(D_s\downarrow\) & Duration$\downarrow$ (s) \\
         \midrule
         BT-H $^{\small\textcolor{gray}{2017}}$            & 0.7098 & 0.1710 & 0.1444 & $\second{2.98}$ \\
         C-BDSD $^{\small\textcolor{gray}{2014}}$        & 0.7475 & 0.2019 & 0.0704 & 104.16 \\
         BDSD-PC$^{\small\textcolor{gray}{2019}}$      & 0.7369 & 0.1723 & 0.1093 & 4.30 \\
         MTF-GLP $^{\small\textcolor{gray}{2006}}$      & 0.7606 & 0.1231 & 0.1330 & 13.08 \\
         MTF-GLP-FS $^{\small\textcolor{gray}{2018}}$ & 0.7680 & 0.1272 & 0.1199 & 9.91 \\
         MF $^{\small\textcolor{gray}{2016}}$              & 0.7709 & 0.1176 & 0.1268 & 7.66 \\
         \midrule
         $\text{CAT}_\text{FusionNet}$          & $\second{0.8554}$ & $\second{0.0832}$ & $\second{0.0680}$ & $\best{2.58}$ \\
         $\text{CAT}_\text{U2Net}$                 & $\best{0.8590}$ & $\second{0.0972}$ & $\best{0.0487}$ & 22.99 \\
         \bottomrule
    \end{tabular}
    }
    \caption{Performance comparison on megapixel images (PAN size of $4000 \times 4000$). The values represent average performance over 4 test images. (\colorbox{red!10}{Red}: best; \colorbox{blue!10}{Blue}: second best)}
    \label{tab:megapixel_small}
\end{table}

As shown in Table \ref{tab:megapixel_small}, most traditional methods perform poorly on megapixel images, and our baseline model also falls short of the top-performing traditional approach. However, with CAT-enhancement, our model achieves state-of-the-art performance across multiple metrics, completing both CAT training and inference in as little as 3 seconds at the fastest setting. \textit{Further details can be found in the supplementary materials.}

\subsection{Limitations and Future Work}
Due to our specific training strategy and loss function design, the proposed module exhibits limited improvements on simulated data generated from simulated data. 
In future work, we plan to optimize the model architecture to enhance its scalability and generalization ability. Additionally, we will explore more effective feature fusion strategies to better exploit the underlying characteristics of simulated data, thereby improving performance on simulated data. These research directions will provide important guidance for further advancing the development of pansharpening techniques.

\section{Conclusions}
This paper proposes an instance-specific approach based on a Conditional Adaptation Tailor (CAT) module, aiming to address the limited generalization capability and low computational efficiency of existing deep learning methods in cross-sensor applications. The innovations of this work lie primarily in two aspects. First, we design a uniquely positioned CAT module that leverages its residual learning capability and strategic placement within the network to enhance the model’s generalization ability. Second, we propose a patch-wise adaptation and parallel inference mechanism to reduce computational overhead. Comprehensive qualitative and quantitative experiments are conducted on especially real-world data across several widely recognized datasets. The results show that our framework outperforms benchmark models composed of several state-of-the-art pansharpening algorithms, achieving SOTA performance across multiple evaluation metrics with rapid speed. Moreover, extensive experiments underscore our framework’s potential to enable effective real-time pansharpening of megapixel data. These findings demonstrate that our module enables rapid adaptation to test data while preserving the performance of pre-trained models, significantly improving the efficiency of deployment and showcasing strong potential for practical applications.

\bibliography{aaai2026}

\clearpage
\appendix

\section{Training Details}

The training process for the proposed model involved a series of carefully chosen hyperparameters to optimize performance while ensuring computational efficiency. We selected 8 patches per image for adaptive training, where each patch was of size $64 \times 64$ in terms of PAN size. The adaptive training was performed over 10 epochs, enabling the model to effectively adapt to the test data.

For the loss function, we utilized a weighted combination of terms, with different weights for real-world and simulated datasets. Specifically, for real-world data, the loss function weights were set as $\eta_1 = 1$, $\eta_2 = 1$, and $\eta_3 = 0.1$. On the other hand, for simulated data, the weights were adjusted to $\eta_1 = 10$, $\eta_2 = 100$, and $\eta_3 = 10000$ to reflect the different scale and characteristics of the data, indicating more conservative adaptation to the synthetic images.

Additionally, for the simulated data adaptation, the final layer of the Conditional Adaptation Tailor (CAT) module was initialized to zero, ensuring that the module starts with a neutral influence before learning any adjustments during the training process.

For optimization, we employed the Adam optimizer \cite{adam}, a robust choice for training deep learning models, with the learning rate set to $1 \times 10^{-4}$ and weight decay at $1 \times 10^{-5}$ to prevent overfitting. The learning rate was carefully tuned to balance convergence speed with stability. We applied the L1 loss criterion for all components of the loss function, which promotes sparsity in the learned features and enhances generalization, particularly when working with high-dimensional data like remote sensing images.

\section{Detailed Discussion}
\subsection{Influence of Number of Selected Patches}
\begin{figure}[!h]
    \centering
    \includegraphics[width=0.8\linewidth]{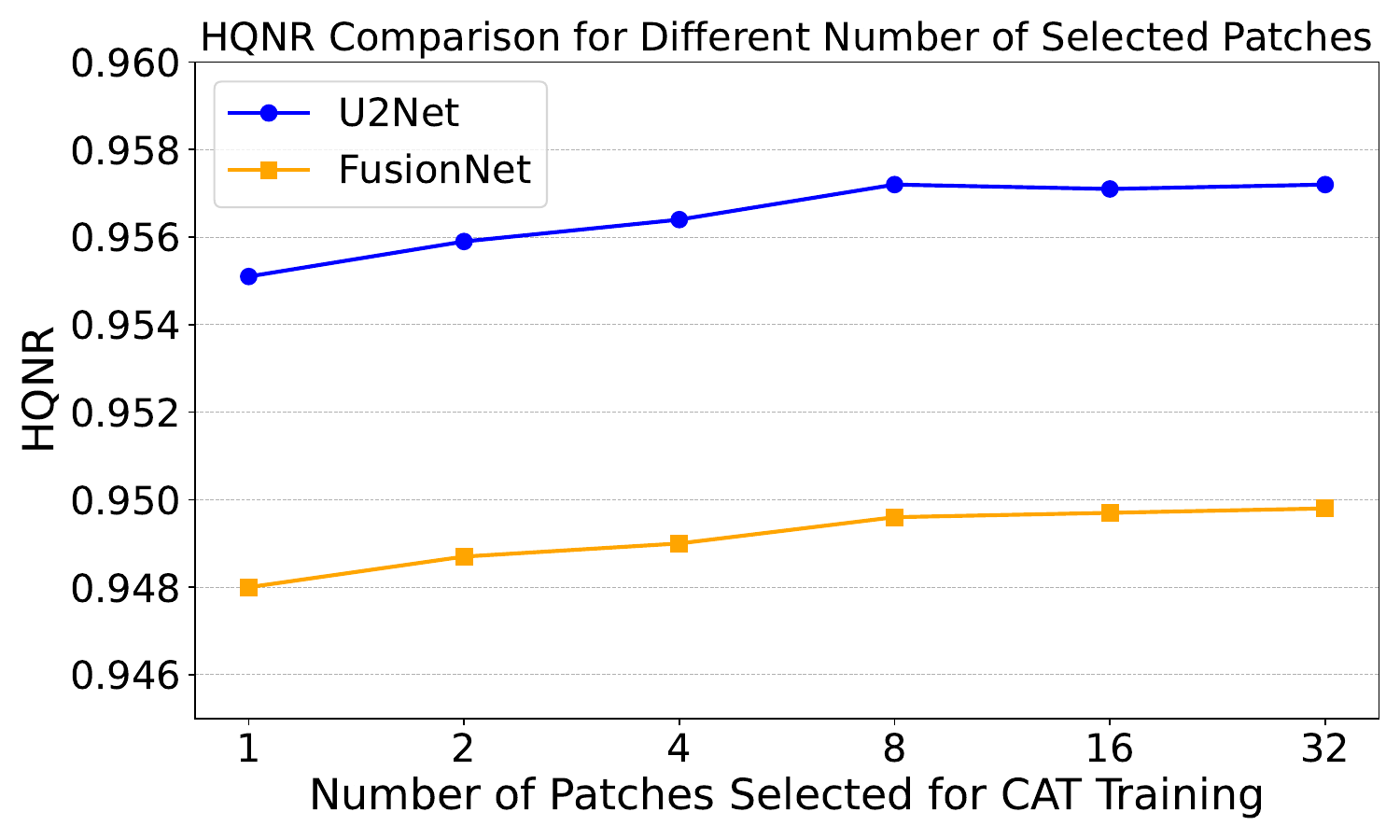}
    \caption{HQNR Metric for U2Net and FusionNet Models with Varying Numbers of Patches on the WV3 Dataset.}
    \label{fig:num_patch}
\end{figure}
In our framework, the number of patches selected for training plays a crucial role in balancing both performance and computational efficiency. While selecting fewer patches reduces the computational load for both training and inference, it also limits the amount of information available for the CAT module to adapt to the test data. 

Therefore we conduct experiments with varying number of selected patches of FusionNet and U2Net on WV3 dataset. We use a $512 \times 512$ PAN image divided into $64\times64$ patches, and we choose part of number from least possible 1 to max possible 64 to ensure a comprehensive analysis.
Here’s the revised version with improved clarity and flow:

As shown in Figure \ref{fig:num_patch}, increasing the number of selected patches initially leads to a rapid improvement in performance. However, as the number of patches continues to rise, the rate of performance improvement gradually slows down. Specifically, this plateau occurs around 8 patches for both models. This suggests that selecting only a small portion—slightly over 10\% of the total patches—provides near-saturated performance, allowing for reduced training time while maintaining high performance. This highlights the efficiency of our framework.

\subsection{Influence of the Patch Size}
\begin{figure}[!h]
    \centering
    \includegraphics[width=\linewidth]{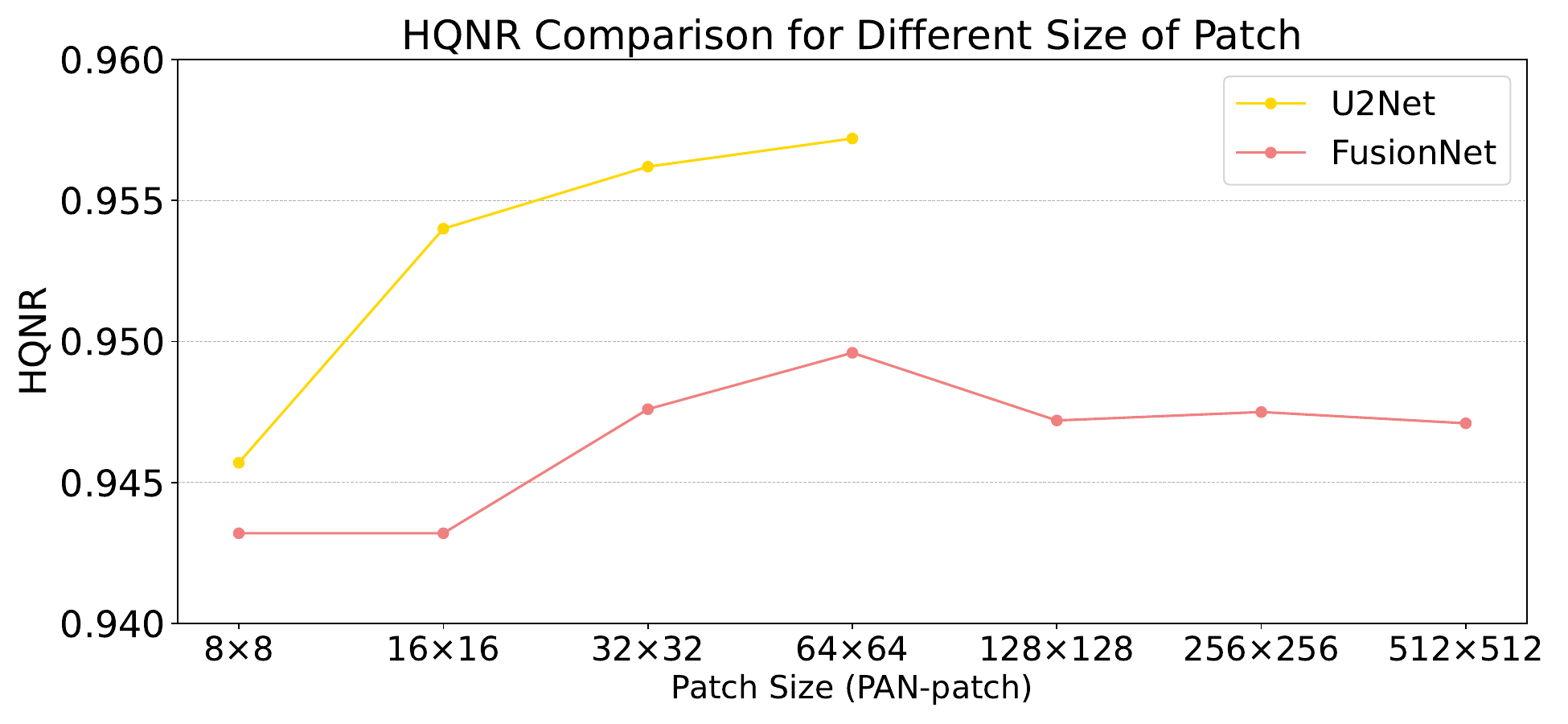}
    \caption{HQNR Metric for U2Net and FusionNet Modelst with Different Patch Sizes on WV3 Dataset.}
    \label{fig:patch_size}
\end{figure}
Patch size is a crucial factor in balancing performance and computational efficiency. Smaller patches reduce computational complexity but provide less information for the CAT module during training, potentially leading to reduced performance. In our experiment, we selected a range of patch sizes within the hardware memory limitations to conduct a comprehensive analysis. The patch sizes ranged from a minimum of $8 \times 8$ to a maximum of $512 \times 512$ in terms of PAN patch size.

As shown in Figure \ref{fig:patch_size}, increasing the patch size initially improves performance for both U2Net and FusionNet, with HQNR rising quickly and peaking at the $64\times64$ size. However, the performance improvement gradually slows as the patch size grows larger. For U2Net, the performance improves consistently up to the largest size that our hardware can tolerate, while FusionNet experiences a slight decline after reaching the peak. 

As the results suggest, while larger patches provide more information and thus enhance performance, using patches as small as $64 \times 64$—significantly smaller than the original $512 \times 512$ PAN image—can already deliver impressive results. This underscores the efficiency and effectiveness of our framework, as it achieves superior performance with reduced computational complexity and faster processing times.

It suggests larger patch with more information is a necessity to achieve better result. Using a patch in the size of $64\times64$-tens of times smaller than the original $512\times512$ PAN-can already generate advanced result, which further speed up the training and inference process, highlighting our efficiency and efficacy. 

\section{Experiment Results on Simulated Data}

Although our method is primarily designed to enhance pansharpening performance on real-world data, with a loss function tailored for real-time applications, it remains effective when applied to simulated data. As shown in Table \ref{tab:reduced_results}, CAT-enhanced networks are able to maintain or slightly improve the performance of their baseline counterparts across multiple evaluation metrics. This demonstrates the generalizability and robustness of the proposed framework, even beyond its original design target.

\begin{table*}[htbp]
  \centering 
  \setlength{\tabcolsep}{4pt}
  \renewcommand\arraystretch{1.2}
  \resizebox{0.95\linewidth}{!}{
    \begin{tabular}{c|cccc|cccc}
      \toprule
      \toprule
      \multirow{2.2}{*}{Method} & \multicolumn{4}{c|}{WV-3 (Simulated Data): Avg$\pm$std} & \multicolumn{4}{c}{WV-2 (Simulated Data): Avg$\pm$std} \\
      \cmidrule{2-5}\cmidrule{6-9}
       & SAM$\downarrow$ & ERGAS$\downarrow$ & sCC$\uparrow$ & Q8$\uparrow$ & SAM$\downarrow$ & ERGAS$\downarrow$ & sCC$\uparrow$ & Q8$\uparrow$ \\
      \midrule
      FusionNet & 3.3494$\pm$0.6310 & 2.4846$\pm$0.6002 & 0.9796$\pm$0.0067 & 0.8954$\pm$0.0913 & 6.3265$\pm$0.6454 & 5.0418$\pm$0.4616 & 0.8813$\pm$0.0136 & 0.7978$\pm$0.0766 \\
      CAT$_{\text{FusionNet}}$   & \second{3.3483$\pm$0.6321} & \second{2.4842$\pm$0.6000} & 0.9796$\pm$0.0067 & \second{0.8957$\pm$0.0909} & \second{6.3255$\pm$0.6457} & 5.0418$\pm$0.4616 & 0.8813$\pm$0.0136 & 0.7978$\pm$0.0766 \\
      \midrule
      U2Net         & 2.8969$\pm$0.5422 & 2.1998$\pm$0.4698 & 0.9858$\pm$0.0041 & 0.9160$\pm$0.0813 & 5.2586$\pm$0.4991 & 4.0764$\pm$0.3832 & 0.9333$\pm$0.0071 & 0.8470$\pm$0.0846 \\
      CAT$_{\text{U2Net}}$       & \second{2.8968$\pm$0.5423} & 2.1998$\pm$0.4698 & 0.9858$\pm$0.0041 & 0.9159$\pm$0.0813 & \second{5.2565$\pm$0.5005} & \second{4.0739$\pm$0.3839} & 0.9333$\pm$0.0071 & \second{0.8472$\pm$0.0845} \\
      \bottomrule
      \bottomrule
    \end{tabular}
  }
  \caption{Performance comparison on simulated data of WorldView-3 (WV-3) and WorldView-2 (WV-2) datasets. The table reports four metrics: SAM, ERGAS, sCC, and Q8, averaged over 20 test images. (\colorbox{blue!10}{Blue}: The improved results compared to the baseline counterpart. }
  \label{tab:reduced_results}
\end{table*}

\section{Application on Megapixel Image}

\begin{figure*}[hbtp]
\centering
\includegraphics[width=0.8\linewidth]{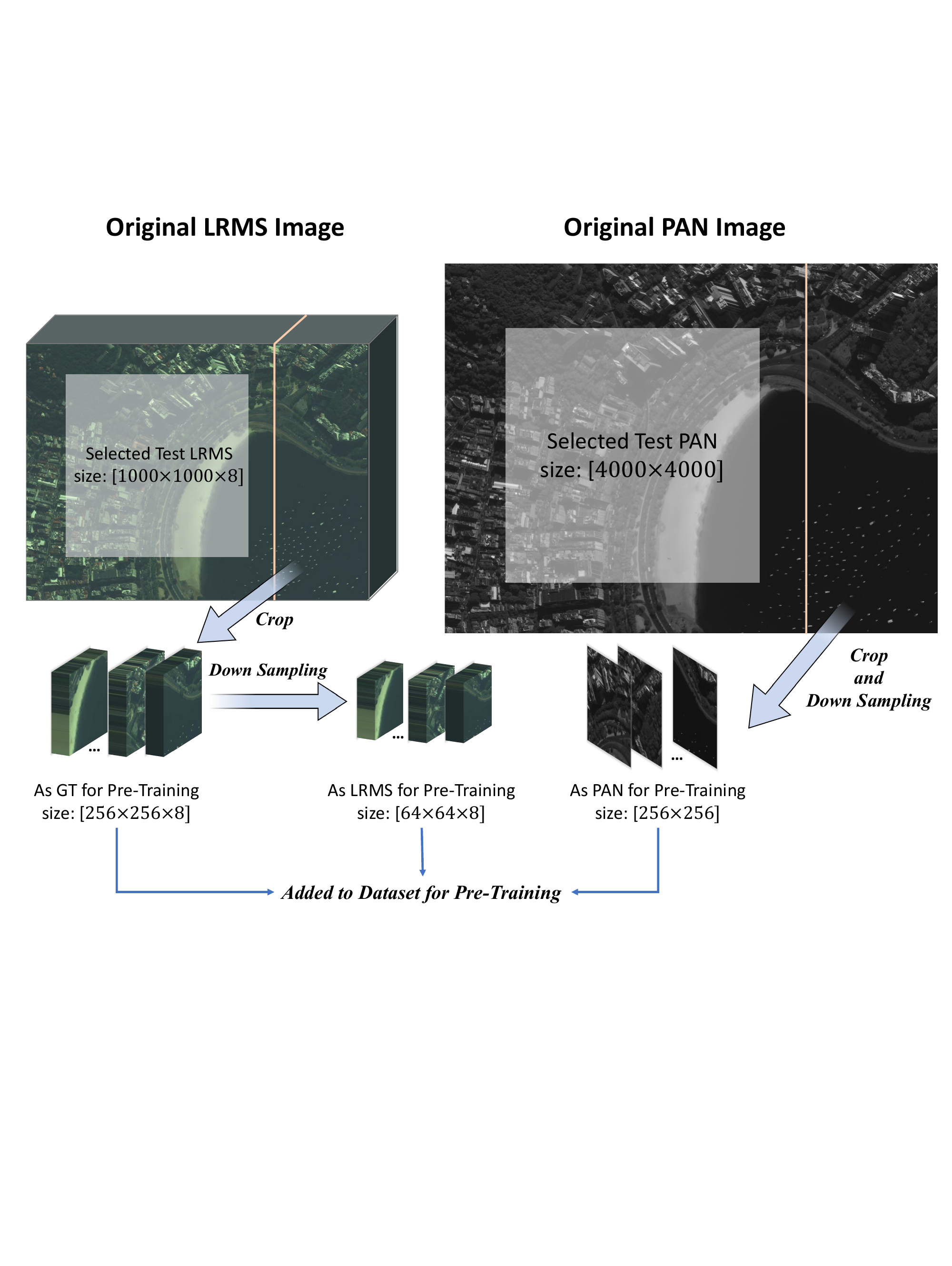}
\caption{Data preparation pipeline for megapixel pansharpening. The pretraining dataset is derived from the right 1/4 portion of the images, while the left 3/4 portion is reserved for test-time evaluation using large-scale megapixel images.}
\label{fig:dataset}
\end{figure*}

The application of deep learning-based pansharpening methods on megapixel images (i.e. images containing more than one million pixels) has long been underexplored, primarily due to the substantial computational overhead and prolonged processing times associated with such large-scale data. Traditional deep learning based methods, while effective on data of smaller images, face challenges when applied to high-resolution satellite imagery, where both computational power and time become limiting factors. 

Our proposed pansharpening method addresses these challenges by enabling efficient adaptation and inference on large-scale images in real time. Unlike traditional deep learning approaches, which suffer from long processing times, our method significantly reduces computational complexity, allowing us to perform both adaptive training and inference on megapixel images in a matter of seconds. This results in high-quality, superior performance fused images, making it a practical solution for real-time applications in megapixel image pansharpening. Our framework thus opens up new possibilities for fast and efficient processing of large-scale satellite data, offering a scalable and highly effective solution for real-world remote sensing tasks on large-scale data.

\subsection{Data Preparation}

Given the lack of an existing dataset specifically designed for megapixel pansharpening, we constructed a custom dataset using four images from WorldView-3, which were utilized both for pretraining the model and for conducting pansharpening evaluation on megapixel-scale images.

For the pretraining phase, we extracted the right 1/4 segment of each image. Specifically, the low-resolution multispectral (LRMS) images were cropped into patches of size $[256 \times 256 \times 8]$, which served as the ground truth (GT) for pretraining. These GT patches were then downsampled to $[64 \times 64 \times 8]$ to serve as the LRMS input data for training. Corresponding panchromatic (PAN) images were cropped into $[1024 \times 1024]$ patches and subsequently downsampled to $[256 \times 256]$ to form the PAN input for pretraining.

For the test-time evaluation, the left 3/4 segment of each image was used. From this portion, we selected a $[1000 \times 1000 \times 8]$ LRMS patch and a corresponding $[4000 \times 4000]$ PAN patch, which were utilized as the test-time input for megapixel image pansharpening.

Using the aforementioned data preparation strategy, we first pre-trained the model on the dataset derived from the right 1/4 segment of the images and subsequently conducted evaluations on the four selected test megapixel images. The results of these experiments are presented in the following subsection.

\subsection{Experiment Results}

To evaluate the efficacy of our method, we conducted experiments on megapixel images, comparing our approach with several existing instance-specific methods. The results which has been demonstrated in the main paper show that traditional methods struggle with efficiency on large-scale data, whereas our method not only shortens processing times to levels comparable to non-deep learning approaches, but also delivers superior performance. This finding demonstrates that deep learning-based pansharpening can be applied efficiently to megapixel images without sacrificing quality, thereby enabling real-time processing of high-resolution imagery.

\begin{figure*}
    \centering
    \includegraphics[width=0.95\linewidth]{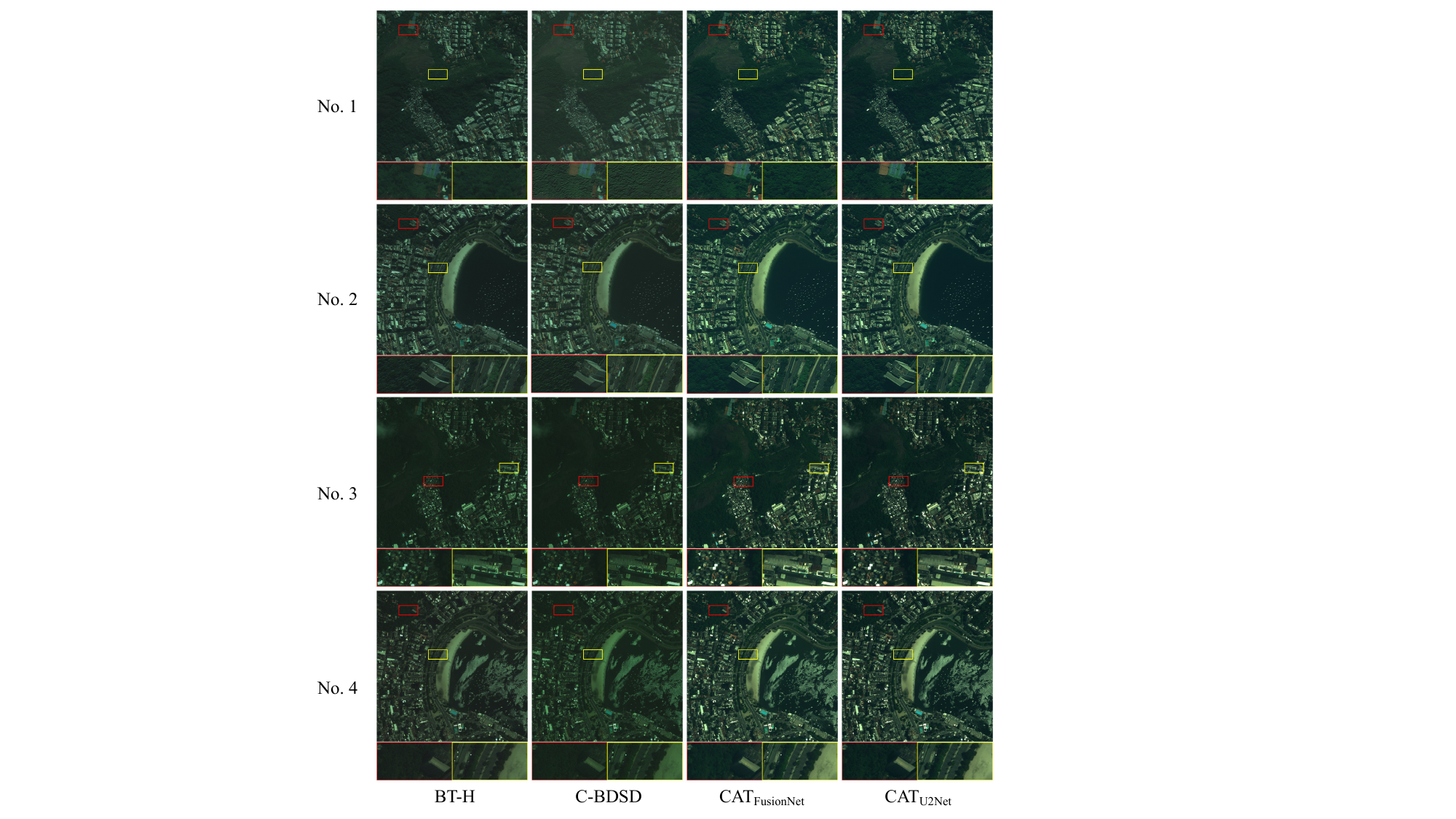}
    \caption{Fusion results on four test images (No. 1–No. 4). Each row corresponds to a different test sample, while each column shows the output from a different method, as labeled below. Red and yellow bounding boxes highlight zoomed-in regions to better illustrate fusion details.}
    \label{fig:big-fig}
\end{figure*}

\begin{figure*}
    \centering
    \includegraphics[width=0.95\linewidth]{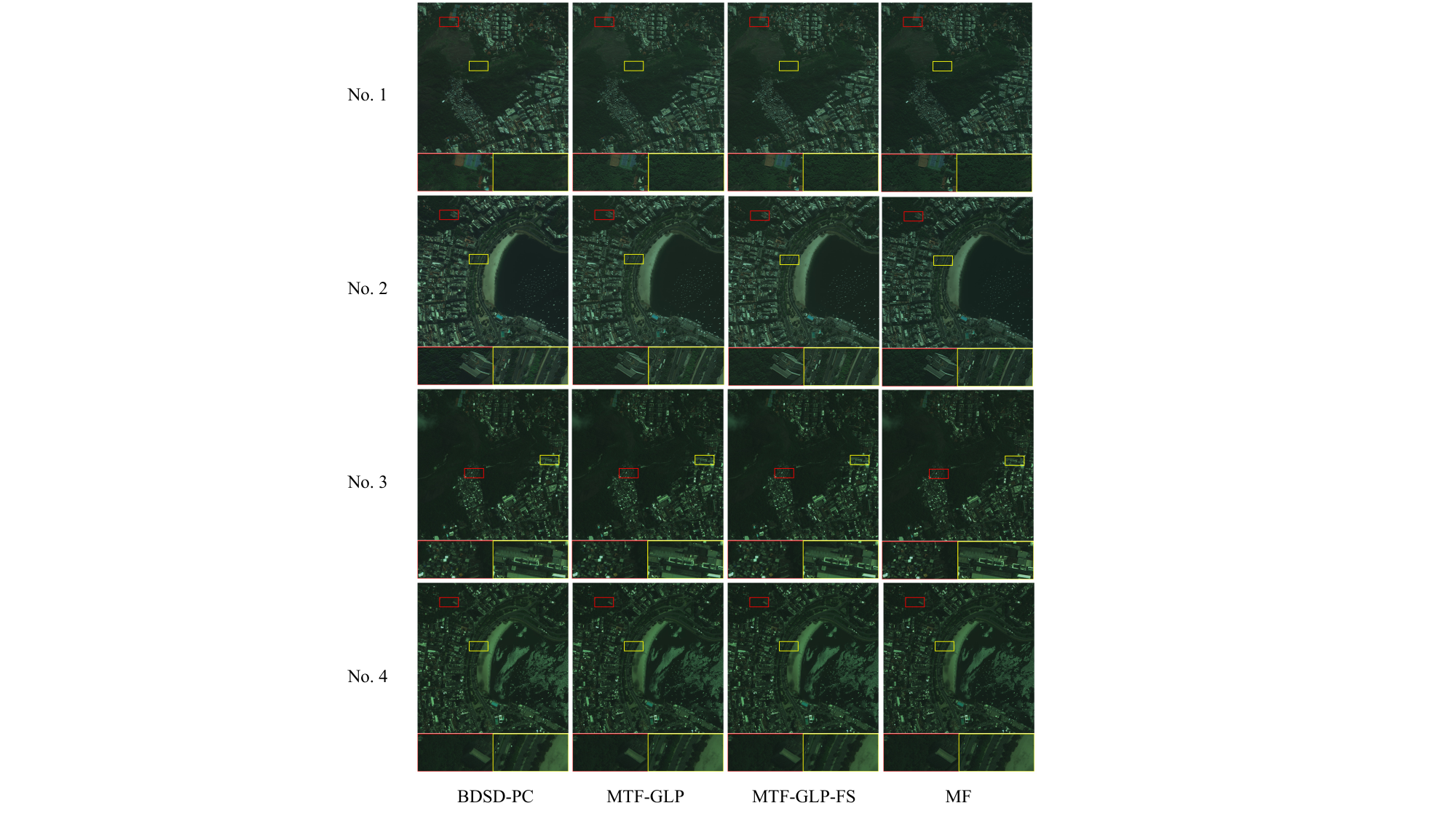}
    \caption{Fusion results on four test images (No. 1–No. 4). Each row corresponds to a different test sample, while each column shows the output from a different method, as labeled below. Red and yellow bounding boxes highlight zoomed-in regions to better illustrate fusion details.}
    \label{fig:big-fig2}
\end{figure*}

Fusion results for four selected test samples obtained using our method, along with the traditional approaches, are presented in Figure \ref{fig:big-fig} and Figure \ref{fig:big-fig2}, where zoomed-in regions are provided to more clearly illustrate the detailed fusion characteristics. Overall, the performance metrics confirm that our approach offers substantial advantages in both the HQNR metric and processing speed, underscoring the practical applicability of our framework for large-scale, real-time pansharpening on megapixel imagery.


\end{document}